\documentclass{article} 
\usepackage{iclr2019_conference,times}

\usepackage{amsmath,amsfonts,bm}

\def\eqref#1{equation~\ref{#1}}

\def\1{\bm{1}}

\DeclareMathAlphabet{\mathsfit}{\encodingdefault}{\sfdefault}{m}{sl}
\SetMathAlphabet{\mathsfit}{bold}{\encodingdefault}{\sfdefault}{bx}{n}
\newcommand{\tens}[1]{\bm{\mathsfit{#1}}}

\def\tB{{\tens{B}}}

\def\tD{{\tens{D}}}

\def\tV{{\tens{V}}}

\usepackage{hyperref}
\usepackage{multirow}
\newcommand{\tabincell}[2]{\begin{tabular}{@{}#1@{}}#2\end{tabular}} 
\usepackage{url}
\usepackage{comment}
\usepackage{graphicx}
\usepackage{wrapfig}
\usepackage[title]{appendix}
\usepackage[ruled, vlined]{algorithm2e}

\makeatletter
\def\thanks#1{\protected@xdef\@thanks{\@thanks
        \protect\footnotetext{#1}}}
\makeatother

\title{Batch Normalization Sampling}

\author{Zhaodong Chen$^{1,2*}$, Lei Deng$^{2*}$, Guoqi Li$^{1\dag}$, Jiawei Sun$^1$, Xing Hu$^2$, Xin Ma$^2$, Yuan Xie$^2$ \\
$^1$Department of Precision Instrument\\
    Center for Brain Inspired Computing Research\\
    Beijing Innovation Center for Future Chip\\
    Tsinghua University\\
$^2$Department of Electrical and Computer Engineering\\
    University of California, Santa Barbara 
    \thanks{$^*$ Indicates equal contribution, $^\dag$ Corresponding to \texttt{liguoqi@mail.tsinghua.edu.cn}}
}

\iclrfinalcopy 
\begin{document}

\maketitle

\begin{abstract}
Deep Neural Networks (DNNs) thrive in recent years in which Batch Normalization (BN) plays an indispensable role. However, it has been observed that BN is costly due to the reduction operations. In this paper, we propose alleviating this problem through sampling only a small fraction of data for normalization at each iteration. Specifically, we model it as a statistical sampling problem and identify that by sampling less correlated data, we can largely reduce the requirement of the number of data for statistics estimation in BN, which directly simplifies the reduction operations. Based on this conclusion, we propose two sampling strategies, ``Batch Sampling" (randomly select several samples from each batch) and ``Feature Sampling" (randomly select a small patch from each feature map of all samples), that take both computational efficiency and sample correlation into consideration. Furthermore, we introduce an extremely simple variant of BN, termed as Virtual Dataset Normalization (VDN), that can normalize the activations well with few synthetical random samples. All the proposed methods are evaluated on various datasets and networks, where an overall training speedup by up to 20\% on GPU is practically achieved without the support of any specialized libraries, and the loss on accuracy and convergence rate are negligible. Finally, we extend our work to the ``micro-batch normalization" problem and yield comparable performance with existing approaches at the case of tiny batch size.
\end{abstract}

\section{Introduction}\label{Sec:Introduction}

Recent years, DNNs have achieved remarkable success in a wide spectrum of domains such as computer vision \citep{krizhevsky2012imagenet} and language modeling \citep{collobert2008unified}. The success of DNNs largely relies on the representation capability benefit from the deep structure \citep{delalleau2011shallow}. However, training a deep network is difficult to converge and batch normalization (BN) has been proposed to solve it \citep{ioffe2015batch}. BN leverages the statistics (mean \& variance) of mini-batches to standardized the activations.(\eqref{equ:BN forward}). It allows the network to go deeper without significant gradient explosion or vanishing \citep{santurkar2018does,ioffe2015batch}. 
Moreover, previous work has demonstrated that BN enables the use of higher learning rate and less awareness on the initialization \citep{ioffe2015batch}, as well as produces mutual information across samples \citep{morcos2018importance} or introduces estimation noises \citep{bjorck2018understanding}  for better generalization. Despite of BN's effectiveness, it is observed that BN introduces considerable training overhead due to the costly reduction operations. The use of BN can lower the overall training speed (training samples per second) by $>$30\%, and the deceleration occurs in both the forward and backward passes \citep{wu2018l1}.

To alleviate this problem, several methods were reported. Range Batch Normalization (RBN) \citep{banner2018scalable} accelerated the forward pass by estimating the variance according to the data range of activations within each batch. A similar approach, $L_1$-norm BN (L1BN) \citep{wu2018l1}, simplified both the forward and backward passes by replacing the $L_2$-norm variance with its $L_1$-norm version and re-derived the gradients for back propagation (BP) training. Different from the above two methods, Self-normalization \citep{klambauer2017self} provided another solution which totally eliminates the need of BN operation with an elaborate activation function called ``scaled exponential linear unit" (SELU). SELU can automatically force the activation towards zero mean and unit variance for better convergence. Nevertheless, all of these methods are not sufficiently effective. The strengths of L1BN \& RBN are very limited since GPU has sufficient resources to optimize the execution speed of complex arithmetic operations such as root in the vanilla calculation of $L_2$-norm variance. Since the derivation of SELU is based on the plain convolutional network, currently it cannot handle other modern structures with skip paths like ResNet and DenseNet.

In this paper, we provide an alternating solution through statistical sampling. Specifically, We demonstrate that randomly sampling a small fraction of activations for the estimation of means and variances before normalization would consistently reduce the computational cost and effectively maintain the convergence rate. Moreover, we demonstrate that it's beneficial to select the less correlated data within each batch. We propose two optimized sampling strategies, ``Batch Sampling" (BS) \& ``Feature Sampling" (FS). BS randomly selects several samples from each batch while FS randomly selects a small patch from each feature map. The estimated statistics (mean \& variance) from the selected small fraction of data can be used to normalize the overall activations well. Inspired by \cite{salimans2016improved}, we further propose an extremely simple variant of BN to balance the conflict of randomness and regularity, termed as Virtual Dataset Normalization (VDN), that uses few synthetical random samples for statistical estimation. All the methods are evaluated on various datasets and networks with different scale (CIFAR-10/CIFAR-100/ImageNet, ResNet/DenseNet). Our experiments demonstrate that up to 20\% speedup for overall training on GPU can be achieved with little accuracy loss. Note that the support of specialized libraries is not needed in our work, which is not like the network pruning \citep{zhu2018structurally} or quantization \citep{hubara2017quantized} requiring extra library for sparse or low-precision computation, respectively. Most previous acceleration work targeted inference which remained the training inefficiency \citep{wen2016learning,molchanov2016pruning,luo2017thinet,zhang2018adam,hu2018novel}, and the rest for training acceleration were orthogonal to our work \citep{lin2017deep,goyal2017accurate,you2017imagenet}. Furthermore, our methods can be extended to the ``micro-batch normalization" (micro-BN) problem and yield advanced training performance with tiny batch size.

In summary, the major contributions of this work are summarized as follows.
\begin{itemize}
    \item We introduce a novel way to alleviate BN's computational cost while maintain the accuracy from the perspective of statistical sampling. Two random sampling strategies are proposed, and an extremely simple variant of BN (i.e. VDN with only few synthetical random samples) is further designed. The VDN can be independently used or combined with any sampling strategy for joint optimization.
    \item Various benchmarks are evaluated, on which up to 20\% practical acceleration for overall training on GPU with negligible accuracy loss and without specialized library support. 
    \item We extend to the micro-BN scenario and achieve advanced performance. 
\end{itemize}

\section{Related Work}\label{Sec:Related}

\textbf{BN} has been applied in most state-of-art models \citep{he2016deep,szegedy2017inception} since it was proposed. As aforementioned, BN standardizes the activation distribution to reduce the internal covariate shift. Models with BN have been demonstrated to converge faster and generalize better \citep{ioffe2015batch,morcos2018importance}. Recently, a model called Decorrelated Batch Normalization (DBN) was introduced which not only standardizes but also whitens the activations with ZCA whitening \citep{2018arXiv180408450H}. Although DBN further improves the normalization performance, it introduces significant extra computational cost.  

\textbf{Simplifying BN} has been proposed to reduce BN's computational complexity. For example, L1BN \citep{wu2018l1} and RBN \citep{banner2018scalable} replace the original $L_2$-norm variance with an $L_1$-norm version and the range of activation values, respectively. From another perspective, Self-normalization uses the customized activation function (SELU) to automatically shift activation's distribution \citep{klambauer2017self}.

However, as mentioned in Introduction, all of these methods fail to obtain a satisfactory balance between the effective normalization and computational cost, especially on large-scale datasets and models. Our work attempts to address this issue.

\textbf{Micro-BN} aims to alleviate the diminishing of BN's effectiveness when the amount of data in each GPU node is too small to provide a reliable estimation of activation statistics. Previous works can be classified to two categories: (1) Sync-BN \citep{zhang2018context} and (2) Local-BN \citep{ba2016layer,wu2018group,ioffe2017batch,wang2018batch}. The former addresses this problem by synchronizing the estimations from different GPUs at each layer, which induces significant inter-GPU data dependency and slows down training process. The latter solves this problem  by either avoiding the use of batch dimension in the batched activation tensor for statistics or using additional information beyond current layer to
calibrate the statistics. Our work can also be extended to tackle with micro-BN problem and achieve advanced performance. 

\section{Problem Formation}
\label{Sec:Problem}
\subsection{Batch Normalization and the Bottleneck Analysis}\label{sec:Bottleneck}

The activations in one layer for normalization can be described by a $d$-dimensional activation feature $\pmb{X} =(x^{(1)},..,x^{(d)})$, where for each feature we have  $x^{(k)} = (x_1^{(k)},..,x_m^{(k)})$. Note that in convolutional (Conv) layer, $d$ is the number of feature maps (FMs) and $m$ equals to the number of points in each FM across all the samples in one batch; while in fully-connected (FC) layer, $d$ and $m$ are the neuron number and batch size, respectively. BN uses the statistics (mean $E[x^{(k)}]$ \& variance $Var[x^{(k)}]$) of the intra-batch data for each feature to normalize that activation by
\begin{equation}
\label{equ:BN forward}
\widehat{x}^{(k)} = \frac{x^{(k)} - E[x^{(k)}]}{\sqrt[]{Var[x^{(k)}]+\epsilon}},~y^{(k)} = \gamma^{(k)}\widehat{x}^{(k)}+\beta^{(k)}
\end{equation}
where $\gamma^{(k)}$ \& $\beta^{(k)}$ are trainable parameters introduced to recover the representation capability, $\epsilon$ is a small constant to avoid numerical error, and $E[x^{(k)}]$ \& $Var[x^{(k)}]$ can be calculated by
\begin{equation}
\label{equ:statistics before sampling}
E[x^{(k)}] = \frac{1}{m}\sum_{j=1}^mx_j^{(k)}, Var[x^{(k)}] = \frac{1}{m}\sum_{j=1}^m(x_j^{(k)}-E[x^{(k)}])^2.
\end{equation}
The detailed operations of a BN layer in the backward pass can be found in Appendix \ref{Sec:Speedup analysis}.
\begin{figure}[h]
\centering
\includegraphics[width=1\textwidth]{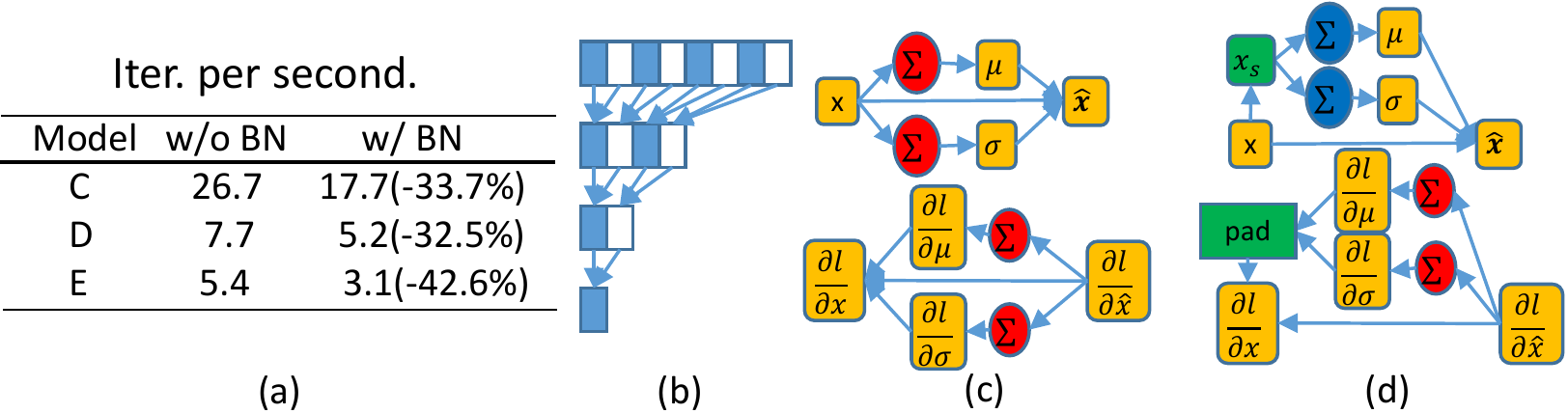}
\caption{\textbf{Illustration of BN cost, operations, and bottleneck}: (a) Iterations per second for training various models with or without BN, Model C,D,E are defined in Table \ref{tab:model configuration}; (b) usual optimization of the reduction operation using adder tree; (c) the computational graph of BN in the forward pass (upper) and backward pass (lower); (d) the computation graph of BN after sampling in forward pass (upper) and backward pass (lower). $x$ is neuronal activations, $\mu$ and $\sigma$ denote the mean and standard deviation of $x$ within one batch, respectively, and $\sum$ is the summation operation.}
\label{fig:Bottleneck}
\end{figure}
From Fig. \ref{fig:Bottleneck}(a), we can see that adding BN will significantly slow down the training speed by 32\%-43\% on ImageNet. The reason of why BN is costly is that it contains several "reduction operations", i.e. $\sum_{j=1}^m$. If the reduction operations are not optimized, it's computational complexity should be $O(m)$. With the optimized parallel algorithm proposed in \cite{che2008accelerating}, the reduction operation is transformed to cascaded adders of depth of $log(m)$ as shown in Fig. \ref{fig:Bottleneck}(b). However, the computational cost is still high since we usually have $m$ larger than one million. Besides, different data in one batch is often generated in different GPU kernels, and this data dependency naturally degrades the parallel benefits. As shown in Fig. \ref{fig:Bottleneck}(c), the red "$\sum$"s represent operations that contain summations, which cause the BN inefficiency.

\subsection{Proposal of BN Sampling}\label{Sec:BN Sampling}
Motivated by above analysis, decreasing the effective value of $m$ at each time for normalization seems a promising way to reduce the BN complexity for achieving acceleration. To this end, we propose BN sampling that samples only a small fraction of data for statistical estimation before normalization at each iteration. In a BN sampling model, Equation (\ref{equ:statistics before sampling}) can be modified as
\begin{equation}
\label{equ:statistics after sampling}
E[x^{(k)}] \approx E[x_s^{(k)}] = \frac{1}{s}\sum_{j=1}^sx_j^{(k)},~ Var[x^{(k)}] \approx Var[x_s^{(k)}] = \frac{1}{s}\sum_{j=1}^s(x_j^{(k)}-E[x_s^{(k)}])^2
\end{equation}
where $x_s^{(k)}$ denotes the sampled data, $s$ is the number of data points after sampling, and we usually have $s \ll m$. We denote \textbf{Sampling Ratio} as $s/m$. The computational graph of BN after sampling is illustrated in Fig. \ref{fig:Bottleneck}(d). The key for BN sampling is how to estimate $E[x^{(k)}]$ \& $Var[x^{(k)}]$ for each neuron or FM within one batch with much less data.

\subsection{Considerations for the Sampling Strategy Design}\label{Sec:Consideration}
We consider the BN sampling from both the perspectives of computational cost and model accuracy. In other words, our target is to simplify BN complexity as low as possible while maintain the accuracy as high as possible. We provide more detailed analysis on the computation cost and accuracy maintaining in Appendix \ref{Sec:Speedup analysis}. The major conclusions are summarized here as below.
\begin{itemize}
    \item Random sampling is promising to estimate the statistics well and can effectively reduce computational cost of BN operations. (Appendix \ref{Sec:Cost} \& Section \ref{sec: acc_eval}) \label{Cl: can speed up }
    
    \item More regular sampling pattern and more static sampling index are expected for practical acceleration. (Appendix \ref{Sec:Cost} \& Section \ref{sec: Accu eval}) \label{cl: static graph}
    
    \item Sampling less correlated data can achieve better estimation in statistics for better accuracy improvement. (Appendix \ref{Sec:Error} \& Section \ref{sec: acc_eval}) \label{cl: correlation}
\end{itemize}

\section{Approaches}\label{Sec:Approach}

In this section, we propose a naive sampling baseline and two random sampling strategies (BS and FS). Furthermore, to balance the conflict of randomness and regularity in BN sampling, we introduce an extremely simple variant of BN (i.e. VDN). Finally, we will describe how we extend our framework to the micro-BN scenario.
\begin{figure}[h]
\centering
\includegraphics[width=1\textwidth]{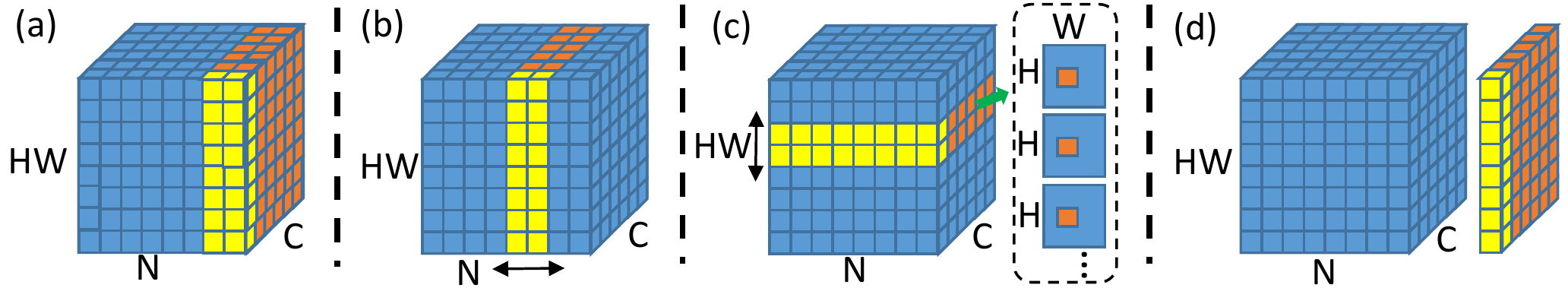}
\caption{\textbf{Illustration of approaches}: (a) Naive Sampling; (b) Batch Sampling; (c) Feature Sampling; (d) Virtual Dataset Normalization. `H' and 'W' are the height and width of FMs, respectively, `C' is the number of FMs for current layer, and `N' denotes the number of samples in current batch. The orange and yellow rectangles in (a)-(c) represent the sampled data and those in in (d) are the virtual sample(s). The yellow data are used to estimate the statistics for current FM.}
\label{fig:Strategies}
\vspace{-15pt}
\end{figure}

\subsection{Sampling Strategies}\label{Sec:Strategies}
In order to design an optimal sampling strategy, it's necessary to jointly consider the aforementioned considerations. In this paper, we propose two sampling strategies: BS and FS. Furthermore, a Naive Sampling (NS) strategy is used for comparison baseline. The sampling strategies are illustrated in Fig. \ref{fig:Strategies}(a)-(c). Specifically: (1) \textbf{NS} fixedly selects $s$ samples to form a reference batch and use its statistics to perform the normalization (Appendix \ref{Sec:algorithm}-Algorithm \ref{Alg:NS BS} in Appendix \ref{Sec:algorithm}).  (2) \textbf{BS} randomly selects $s$ samples to form a reference batch (Appendix \ref{Sec:algorithm}-Algorithm \ref{Alg:NS BS}).  (3) \textbf{FS} randomly selects a small patch from each feature map (FM) across all samples (Appendix \ref{Sec:algorithm}-Algorithm \ref{Alg:FS}).

\textbf{Data Randomness and Correlation}. Here we make two empirical assumptions: within each layer or between layers, (1) data in the same sample are more likely to be correlated; (2) data sharing the same location in FMs are more likely to be correlated. Under these assumptions, BS or FS recovers BN's effectiveness by randomly selecting sample indexes or patch indexes, respectively, to reduce the data correlation as much as possible. In the statistical sense, they are able to estimate the overall information. Specifically, the variable sample indexes (BS) or patch indexes (FS) across different layers guarantee less correlation in data for normalization. Since FS's sampled data come from more samples compared with NS and BS, it brings better statistical estimation.

\textbf{Sampling Pattern and Index}. Besides the randomness for statistical estimation, according to the second consideration in Section \ref{Sec:Consideration}, we expect more regular sampling pattern and more static sampling index for practical acceleration. Therefore, to balance the estimation loss and execution acceleration, we give the following sampling rules: (1) In NS, the sample index for different channels in each layer and different layers are shared; (2) In BS, the selected samples are continuous and the sample index for different channels are shared, while they are independent between different layers; (3) In FS, the patch shape is rectangular and the patch location is shared by different channels and samples within each layer but variable as layer changes. Furthermore, all the random indexes are updated only once for each epoch.

\subsection{Virtual Dataset Normalization}\label{Sec:VDN}

In fact, there is an implicit conflict between the last two considerations in Section \ref{Sec:Consideration}. The requirements for more regular sampling pattern and static index would improve the efficiency of memory access and the related computation, however, it will increase the data correlation resulting in degraded normalization quality. To balance this conflict, we propose an extremely simple variant of BN, named VDN that completes the normalization using only one synthetical sample. VDN can be implemented with the following three steps: (1) calculating the statistics of the whole training dataset offline; (2) generating $s$ random inputs (virtual samples) at each iteration to concatenate with the original real inputs as the final network inputs; (3) using data from either only virtual samples or the combination of virtual and real samples at each layer for statistical estimation, where the real data can be sampled by any sampling strategies mentioned in Section \ref{Sec:Strategies}. VDN depicted in Fig. \ref{fig:Strategies}(d) and the detailed algorithm is given in Appendix \ref{Sec:algorithm}-Algorithm \ref{Alg:VDN}. Combining VDN with other sampling strategy can be described as
\begin{equation}
\label{equ:vdn+}
X[x^{(k)}] = \beta X[x_v^{(k)}] + (1-\beta)X[x_s^{(k)}]
\end{equation}
where $X$ stands for ``$E$'' or ``$Var$'', $x_v$ is the virtual data while $x_s$ represents the sampled data. $\beta$ is a controlling variable: (1) when $\beta=$ 0 or 1, the statistics come from single method (VDN or any sampling strategy); (2) when $\beta\in (0,~1)$, the final statistics are a joint value. An optimal $\beta$ may be obtained through extensive cross-validation, whereas in this paper, we set $\beta=0.5$ for simplicity. VDN shares some similar attributes with previous work named Virtual Batch Normalization (VBN) \citep{salimans2016improved}. VBN was proposed to avoid the problem in which an input $x$ is highly dependent on some of other inputs within the same mini-batch in generative adversarial networks (GANs). In VBN, each sample $x$ is normalized based on the statistics collected from a reference batch of samples. The reference batch is as large as the real one so it introduces considerable computation cost, which is the reason why VBN is only used in generator. In our VDN, we extend to general tasks like image recognition and experiments will demonstrate that the randomly-generated virtual samples are more effective compared to the real ones.

\subsection{Micro-BN}
We further extend our framework to the micro-BN scenario, which also faces the problem that the amount of data used for statistics is too small. In Sync-BN: (1) \textbf{With FS}, each GPU node executes the same patch sampling as normal FS; (2) \textbf{With BS}, we can randomly select the statistics from a fraction of GPUs rather than all nodes; (3) \textbf{With VDN}, the virtual samples can be fed into a single or few GPUs. The first one can just simplify the computational cost within each GPU, while the last two further optimize the inter-GPU data dependency. In Local-BN, since the available data for each GPU is already tiny, the BN sampling strategy will be invalid. Fortunately, the VDN can still be effective by feeding virtue samples into each node.

\vspace{20pt}
\section{Experiments}
\begin{wraptable}[9]{r}{0.6\textwidth}
\caption{Model configuration.}
\centering
\resizebox{0.6\textwidth}{!}{
\begin{tabular}{ccccc}
\hline
\bf Model  &\bf Dataset		&\bf Network &\bf Samples/GPU &\multicolumn{1}{c}{\bf GPU} 
\\ \hline 
A &CIFAR-10 &ResNet-20 & 128 & Titan XP $\times 1$\\
B &CIFAR-10/100 &ResNet-56 & 128 & Titan XP $\times 1$\\
C &ImageNet &ResNet-18 & 32 & V100 $\times 1$\\
D &ImageNet &ResNet-18 & 128 & V100 $\times 2$\\
E &ImageNet &ResNet-50 & 64 & V100 $\times 4$\\
F &ImageNet &DenseNet-121 & 64 & V100 $\times 3$\\
\hline
\end{tabular}}
\label{tab:model configuration}
\end{wraptable}
\textbf{Experimental Setup}. All of our proposed approaches are validated on image classification task using CIFAR-10, CIFAR-100 and ImageNet datasets from two perspective: (1) Accuracy Evaluation and (2) Acceleration Evaluation. To demonstrate the scalability and generality of our approaches on deep networks, we select ResNet-56 on CIFAR-10 \& CIFAR-100 and select ResNet-18 and DenseNet-121 on ImageNet. The model configuration can be found in Table \ref{tab:model configuration}. The means and variances for BN are locally calculated in each GPU without inter-GPU synchronization as usual. We denote our approaches as the format of ``Approach-Sampled\_size/Original\_size-sampling\_ratio(\%)". For instance, if we assume batch size is 128, ``BS-4/128-3.1\%" denotes only 4 samples are sampled in BS and the sampling ratio equals to $\frac{4}{128}=3.1\%$. Similarly, ``FS-1/32-3.1\%" implies a $\frac{1}{32}=3.1\%$ patch is sampled from each FM, and ``VDN-1/128-0.8\%"  indicates only one virtual sample is added. The traditional BN is denoted as ``BN-128/128-100.0\%". Other experimental configurations can be found in Appendix \ref{Sec:exp config}.

\subsection{Accuracy Evaluation}\label{sec: Accu eval}
\begin{figure}[htbp]
\centering
\includegraphics[width=0.8\textwidth]{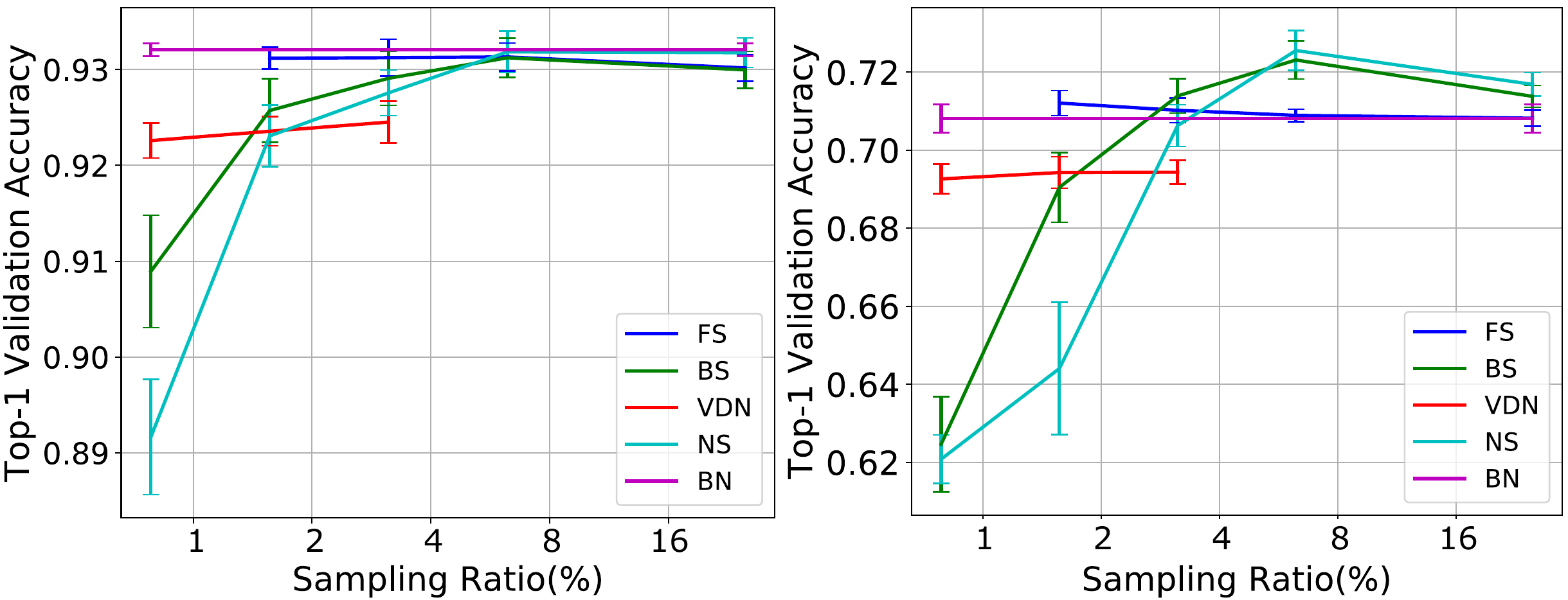}
\caption{Top-1 validation accuracy of ResNet-56 on CIFAR-10 (left) \& CIFAR-100 (right).}
\label{fig:accuracy cifar10}
\end{figure}

\begin{figure}[htbp]
\centering
\includegraphics[width=0.83\textwidth]{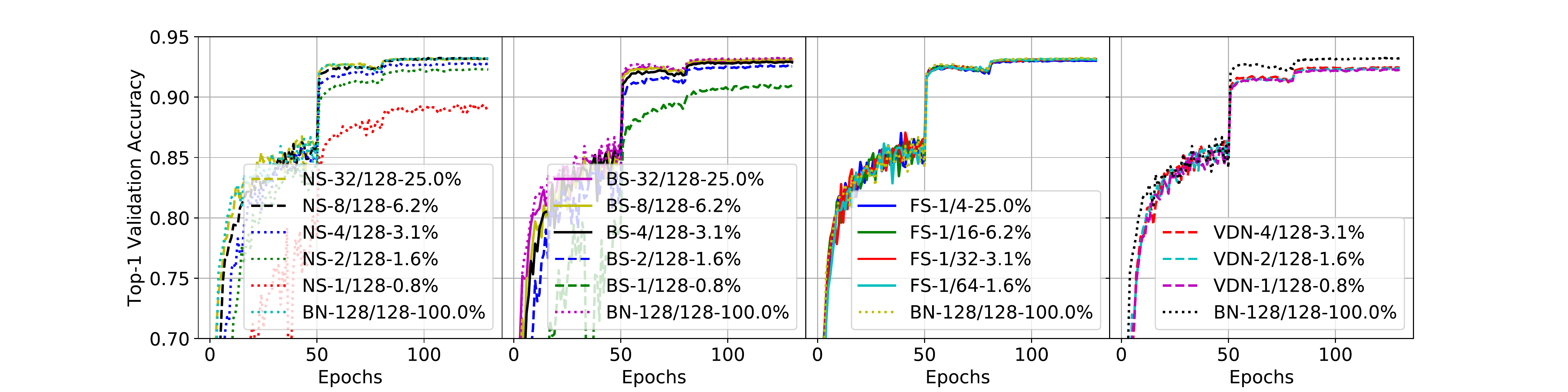}
\caption{Training curves of ResNet56 on CIFAR-10.}
\label{fig:training curve cifar10}
\vspace{-12pt}
\end{figure}

\textbf{Convergence Analysis}. Fig. \ref{fig:accuracy cifar10} shows the top-1 validation accuracy and confidential interval of ResNet-56 on CIFAR-10 and CIFAR-100. On one side, all of our approaches can well approximate the accuracy of normal BN when the sampling ratio is larger than $2\%$, which evidence their effectiveness. On the other side, all the proposed approaches perform better than the NS baseline. In particular, \textbf{FS} performs best, which is robust to the sampling ratio with negligible accuracy loss (e.g. at sampling ratio=1.6\%, the accuracy change is \textbf{-0.0871\%} on CIFAR-10 and \textbf{+0.396\%} on CIFAR-100). 
\textbf{VDN} outperforms BS and NS with a large margin in extremely small sampling ratio (e.g. 0.8\%), whereas the increase of virtual batch size leads to little improvement on accuracy. \textbf{BS} is constantly better than NS. Furthermore, an interesting observation is that the BN sampling could even achieve better accuracy sometimes, such as NS-8/128\textbf{(72.6$\pm$1.5\%)}, BS-8/128\textbf{(72.3$\pm$1.5\%)}, and FS-1/64\textbf{(71.2$\pm$0.96\%)} against the baseline \textbf{(70.8$\pm$1\%)} on CIFAR-100. 
Fig.\ref{fig:training curve cifar10} further shows the training curves of ResNet-56 on CIFAR-10 under different approaches.  It reveals that FS and VDN would not harm the convergence rate, while BS and NS begin to degrade the convergence when the sampling ratio is smaller than 1.6\% and 3.1\%, respectively. 

\begin{table}[t]
\caption{Top-1 validation error on ImageNet.}
\centering
\resizebox{0.8\textwidth}{!}{
\begin{tabular}{c|c|c|c|c}
\hline
{\bf Model}& {\bf Approach} &{\bf Sampling Ratio} &{\bf Top-1 Error(\%)} &{\bf Accuracy Loss(\%)} 
\\ \hline \hline
\multirow{9}{*}{\tabincell{c}{\textbf{ResNet-18}\\(256, 128)}} & BN & 128/128(100\%) & 29.8 & baseline\\
\cline{2-5}
&\multirow{2}{*}{NS} &1/128(0.78\%) & N.A. &N.A.\\
 &&4/128(3.1\%) &35.2 & -5.42\\
\cline{2-5}
&\multirow{2}*{BS} &1/128(0.78\%) & N.A. & N.A.\\
 &&4/128(3.1\%) &31.7 & -1.9\\
\cline{2-5}
&\multirow{2}*{VDN} &1/128(0.78\%) &31.2 &-1.4\\
 &&2/128(1.6\%) &30.8 & -1.0\\
\cline{2-5}
&\multirow{2}*{FS} &1/64(1.6\%) &30.3 &-0.5\\
& &1/32(3.1\%) &30.4 &-0.6\\
\cline{2-5}
&\textbf{FS+VDN} &4/128(3.1\%) &30.0 & \textbf{-0.2}\\
\hline
\hline
\multirow{2}{*}{\tabincell{c}{\textbf{DenseNet-121}\\(192, 64)}} &BN & 64/64(100\%) &26.1 & baseline\\
\cline{2-5}
&\textbf{FS+VDN} & 3/64(4.7\%) &26.7&\textbf{-0.6}\\
\hline
\end{tabular}}
\label{tab:imagenet error}
\vspace{-10pt}
\end{table}

\begin{figure}[!htbp]
\centering
\includegraphics[width=0.8\textwidth]{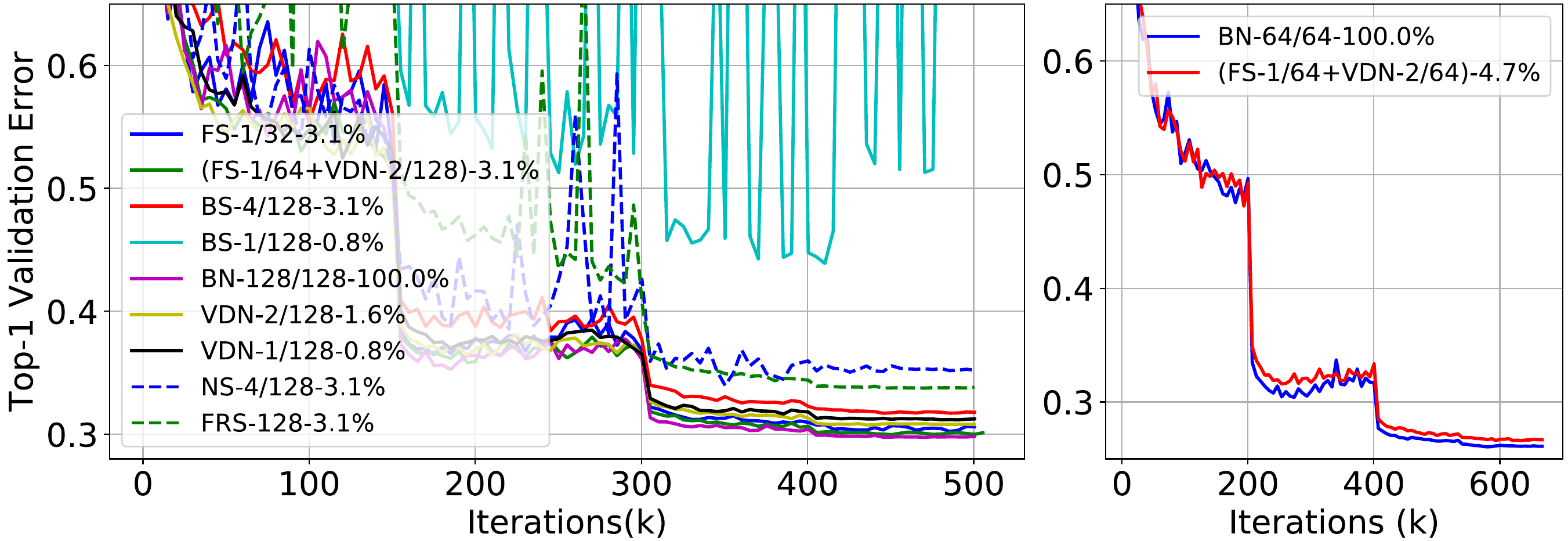}
\caption{Training curves of ResNet18 (left) and DenseNet121 (right) on ImageNet.}
\label{fig:training curve imagenet}
\vspace{-12pt}
\end{figure}

Table \ref{tab:imagenet error} shows the top-1 validation error on ImageNet under different approaches. With the same sampling ratio, all the proposed approaches significantly outperform NS, and \textbf{FS} surpasses VDN and BS. Under the extreme sampling ratio of 0.78\%, NS and BS don't converge. Due to the limitation of FM size, the smallest sampling ratio of \textbf{FS} is 1.6\%, which has only -0.5\% accuracy loss. \textbf{VDN} can still achieve relatively low accuracy loss (1.4\%) even if the sampling ratio decreases to 0.78\%. This implies that VDN is effective for normalization. Moreover, by combining FS-1/64 and VDN-2/128, we get the lowest accuracy loss (-0.2\%). This further indicates that VDN can be combined with other sampling strategies to achieve better results. Since training DenseNet-121 is time-consuming, we just report the results with \textbf{FS-VDN mixed} approach. Although DenseNet-121 is more challenging than ResNet-18 due to the much deeper structure, the ``FS-1/64 + VDN-2/64" can still achieve very low accuracy loss (-0.6\%). In fact, we observed gradient explosion if we just use VDN on very deep network (i.e. DenseNet-121), which can be  conquered through jointly applying VDN and other proposed sampling strategy (e.g. FS+VDN). Fig. \ref{fig:training curve imagenet} illustrates the training curves for better visualization of the convergence. Except the BS with extremely small sampling ratio (0.8\%) and NS, other approaches and configurations can achieve satisfactory convergence. Here, we further evaluate the fully random sampling (FRS) strategy, which samples completely random points in both the batch and FM dimensions. We can see that FRS is less stable compared with our proposed approaches (except the NS baseline) and achieves much lower accuracy. One possible reason is that under low sampling ratio, the sampled data may occasionally fall into the worse points, which lead to inaccurate estimation of the statistics.

\textbf{Correlation Analysis}.\label{part: corr_analyze} In this section, we bring more empirical analyze on the data correlation that affects the error of statistical estimation. Here we denote the estimation errors as $E_{\mu}^{(i)} = ||\mu_s^{(i)} - \mu^{(i)}||_2$ and $E_{\sigma}^{(i)} = ||\sigma_s^{(i)} - \sigma^{(i)}||_2$, where $\mu_s^{(i)}$ \& $\sigma_s^{(i)}$ are the estimated mean \& variance from the sampled data while $\mu^{(i)}$ \& $\sigma^{(i)}$ are the ground truth for the whole batch.

\begin{figure}[!htbp]
\vspace{5pt}
\centering
\includegraphics[width=0.9\textwidth]{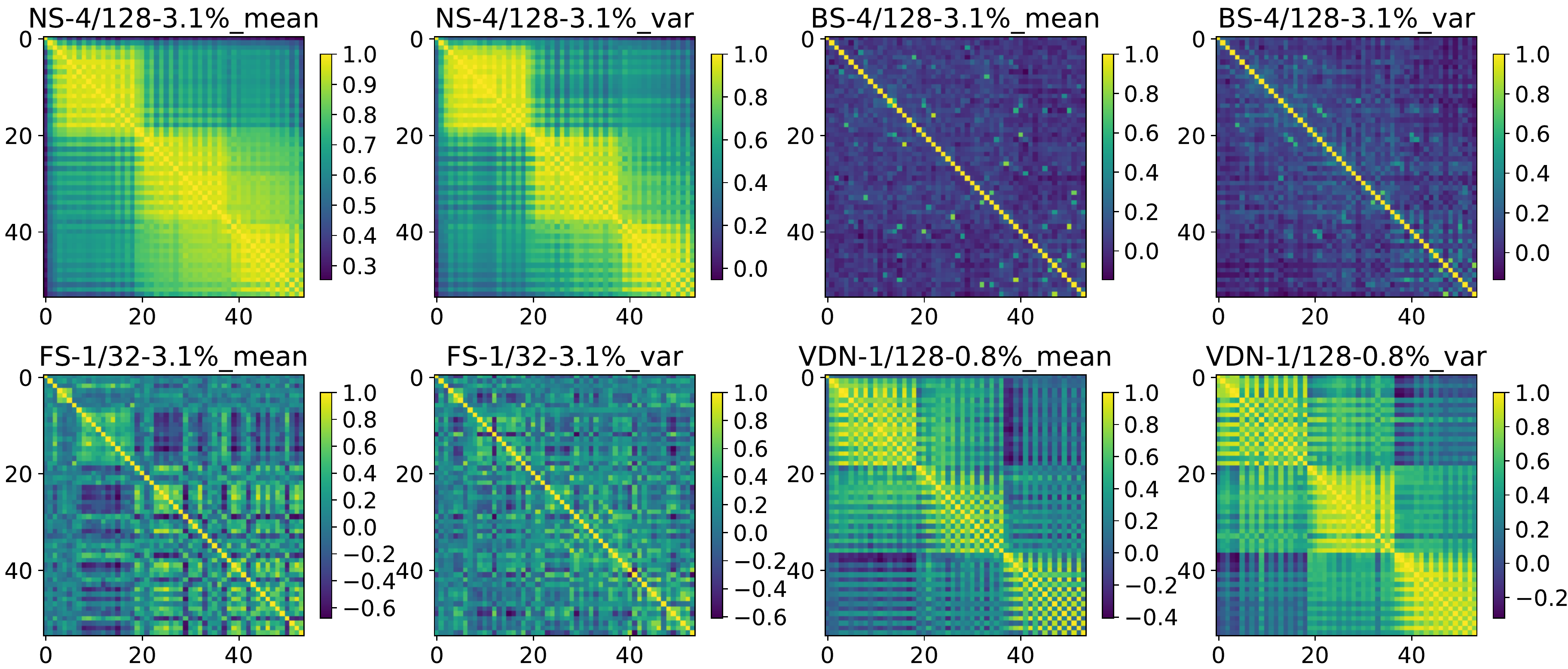}
\caption{Inter-layer correlation between estimation errors.}
\label{fig:correlation inter layers}
\vspace{-10pt}
\end{figure}

\begin{wrapfigure}[14]{l}{0.45\textwidth}
\includegraphics[width=0.45\textwidth]{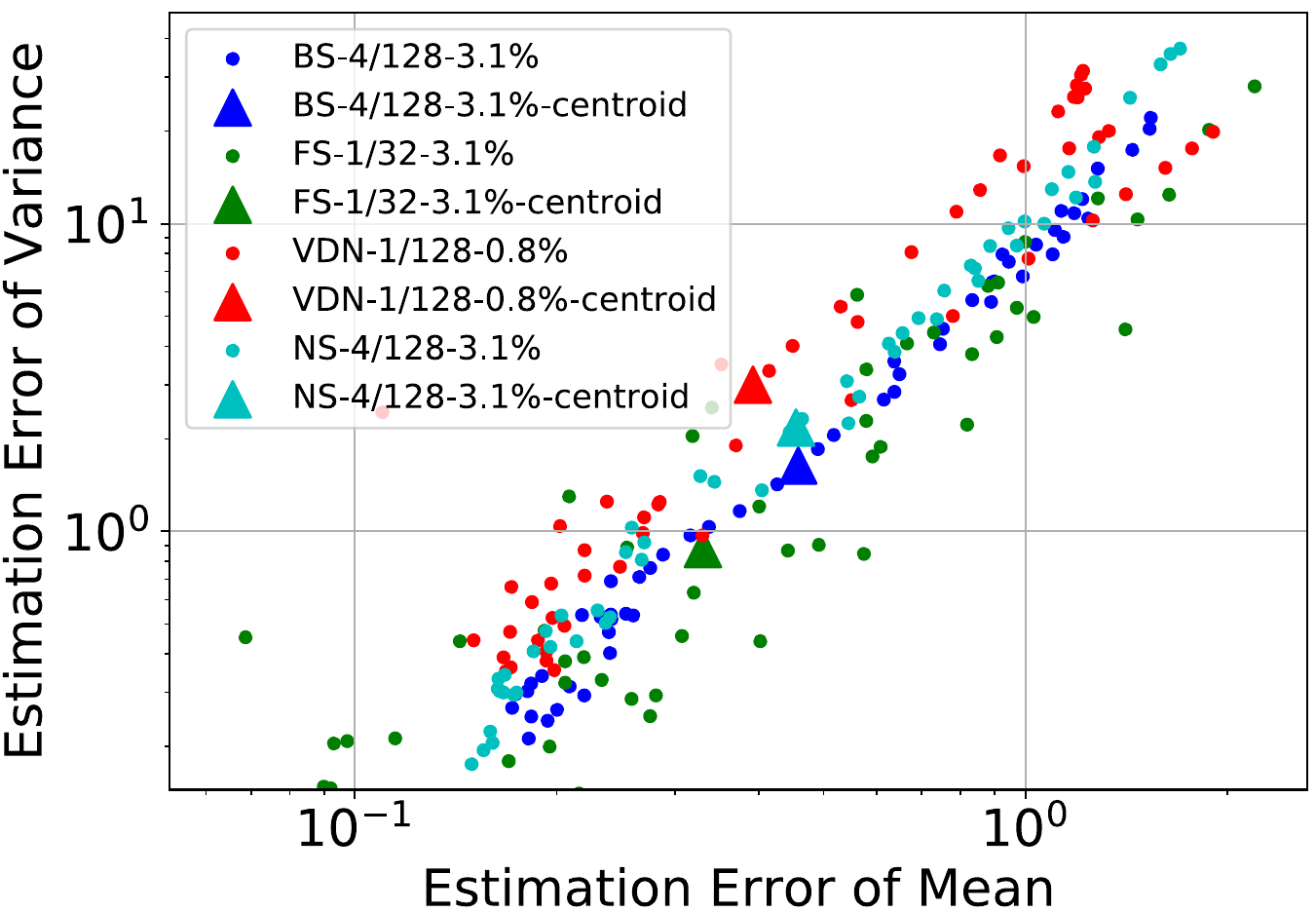}
\caption{Estimation error distribution.}
\label{fig:intra-layer estimation}
\vspace{-8pt}
\end{wrapfigure}

The analysis is conducted on ResNet-56 over CIFAR-10. The estimation errors of all layers are recorded throughout the first epoch. Fig. \ref{fig:correlation inter layers} and Fig. \ref{fig:intra-layer estimation} present the inter-layer correlation between estimation errors and the distribution of estimation errors for all layers, respectively. From Fig. \ref{fig:correlation inter layers}, we can see that BS presents obviously less inter-layer correlation than NS, which is consistent with previous experimental result that BS converges better than NS. 
For FS and VDN, although it looks like they have averagely higher correlations, there exists negative corrections which effectively improves the model accuracy. Moreover, FS produces better accuracy than NS and BS since its selected data come from all samples with less correlation (details in Appendix \ref{Sec:Error}). In Fig. \ref{fig:intra-layer estimation}, FS also shows the least estimation error which further implies its better convergence. The estimation error of VND seems similar to BS and NS here, but we should note that it uses much lower sampling ratio of 0.8\% compared to others (3.1\%). Although BS and NS have similar estimation error, BS can converge better since it has less inter-layer correlation shown in Fig. \ref{fig:correlation inter layers}.

\subsection{Acceleration Evaluation}\label{sec: acc_eval}

\begin{figure}[htbp]
\centering
\includegraphics[width=1\textwidth]{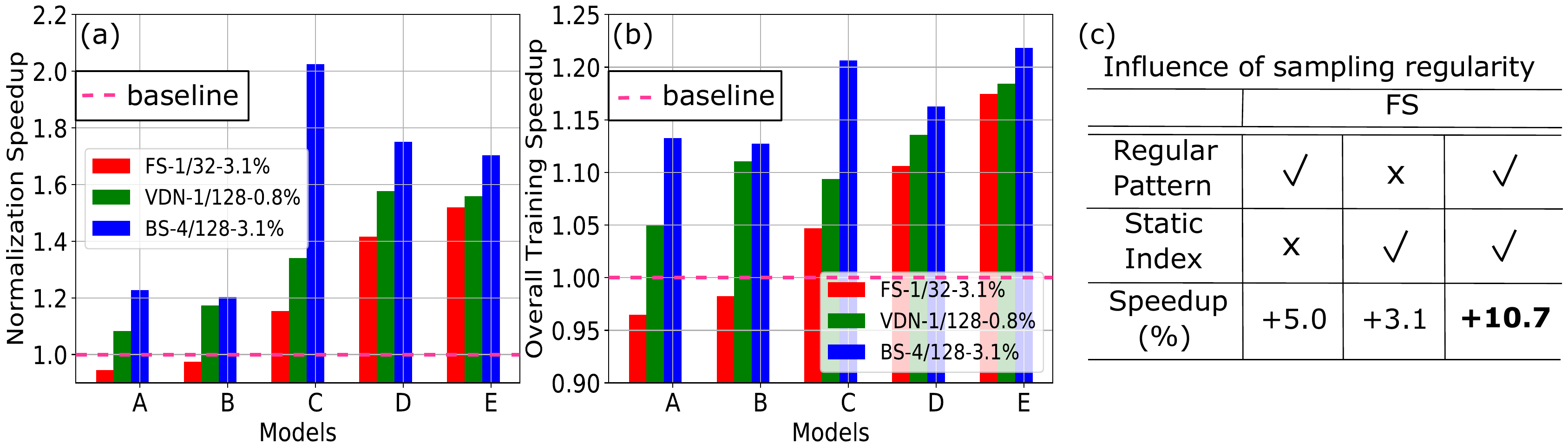}
\caption{Acceleration evaluation: (a) normalization speedup; (b) overall training speedup; (c) influence of sampling regularity (ImageNet, ResNet-18, FS-1/32-3.1\%).}
\label{fig:speedup}
\vspace{-15pt}
\end{figure}

After the accuracy evaluation, we will evaluate the acceleration benefit which is our motivation. Fig. \ref{fig:speedup} shows the normalization speedup during training and overall training improvement. 

In general, BS can gain higher acceleration ratio because it doesn't incur the fine-grained sampling within FMs like in FS and it doesn't require the additional calculation and concatenation of the virtual samples like in VDN. As for FS, it fails to achieve speedup on CIFAR-10 due to the small image size that makes the reduction of operations unable to cover the sampling overhead. The proposed approaches can obtain up to 2x BN acceleration and 20\% overall training acceleration. Furthermore, Table \ref{tab:lateral comparison} gives additional results on DenseNet-121 and provides lateral comparisons with other methods for BN simplification. Our approaches perform faster training compared with two recent methods. On very deep networks with more BN layers, such as DenseNet-121, the speedup is more significant. It's worthy noting that, our training speedup is obtained without the support of specialized library that makes it easy-to-use. Fig. \ref{fig:speedup} (c) reveals that "Regular pattern" and "Static Index" help us achieve practical speedup.
\begin{table}[htbp]
\vspace{-15pt}
\caption{Lateral comparison of overall training speed.}
\centering
\resizebox{0.85\textwidth}{!}{
\begin{tabular}{c|c|c|c|c}
\hline
{\bf Approach}  &{\bf Model}		&{\bf Sampling Ratio} &{\bf Iter. per Second} &{\bf Speedup(\%)} 
\\ \hline \hline
\multirow{5}{*}{\tabincell{c}{\textbf{ResNet-18}\\(256, 128)}}&BN&128/128 & 5.21 & baseline\\
\cline{2-5}
&L1BN\citep{wu2018l1}& 128/128 &5.23 & +0.38\\
\cline{2-5}
&RBN\citep{banner2018scalable}& 128/128 &5.30 &+1.73 \\
\cline{2-5}
&FS&1/32(3.1\%) & 5.77 & +10.7\\
\cline{2-5}
&VDN&1/128(0.78\%) & 5.93 & +13.8\\
\cline{2-5}
&BS&4/128(3.1\%) & 6.07 & +16.5\\
\hline \hline
\multirow{2}{*}{\tabincell{c}{\textbf{DenseNet-121}\\(192, 64)}} &BN&64/64 & 2.44 & baseline\\
\cline{2-5}
&FS+VDN & 3/64(4.7\%) & 2.97 & +21.7\\
\hline
\end{tabular}}
\label{tab:lateral comparison}
\vspace{-15pt}
\end{table}

\subsection{Micro-BN Extension}
\begin{wrapfigure}{r}{0.28\textwidth}
\includegraphics[width=0.28\textwidth]{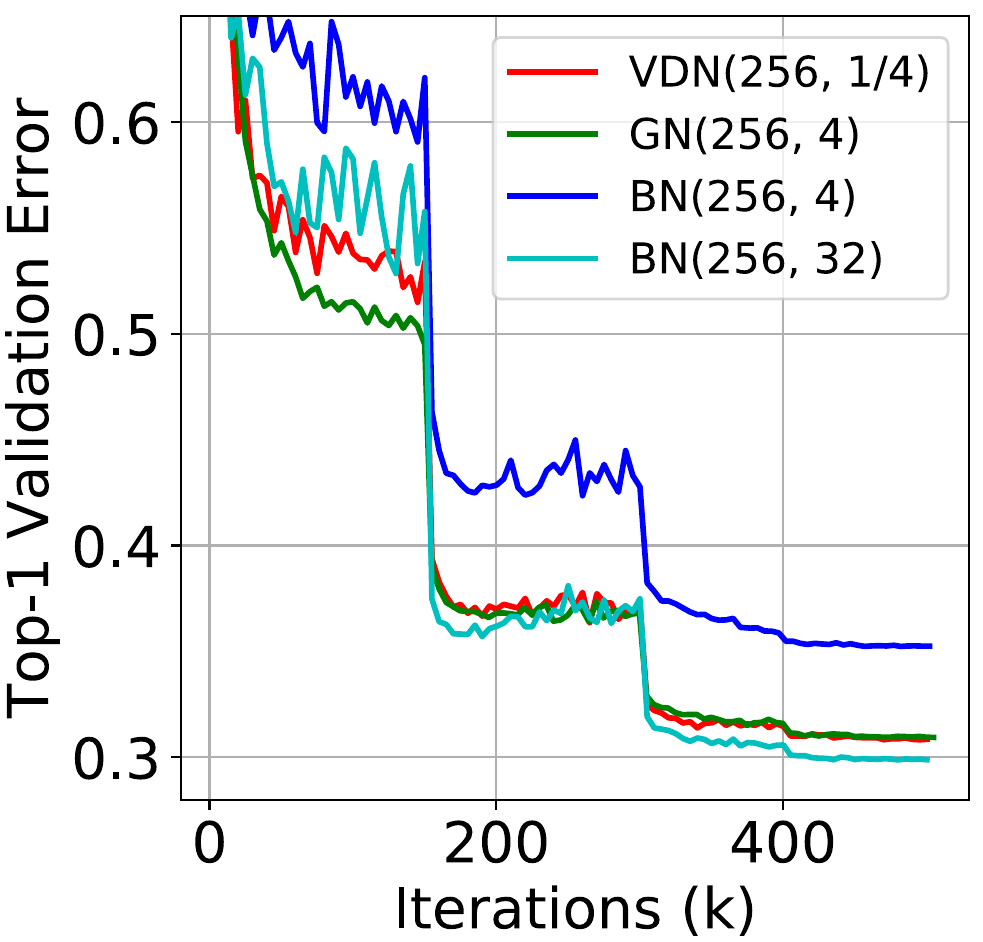}
\caption{Training curves in micro-BN case. (ImageNet, ResNet 18)}
\label{fig:micro-BN}
\end{wrapfigure}
At last we extend our work to the micro-BN case where the batch size is significantly reduced on each GPU node. The normalization in Sync-BN is based on the statistics from multiple nodes through synchronization, which is equivalent to that in Fig. \ref{fig:training curve imagenet} with large batch size for each node. Therefore, to avoid repetition, here we just show the results on Local-BN with VDN optimization. We let the overall batch size of 256 break down to 64 workers (each one has only 4 samples for local normalization). We use ``(gradient batch size, statistic batch size)" of (256, 4) to denote the configuration \citep{wang2018batch}. A baseline of (256, 32) with BN and one previous work Group Normalization (GN) \citep{wu2018group} are used for comparison. As shown in Fig. \ref{fig:micro-BN}, although the reduction of batch size will degrade the model accuracy, our VDN can achieve slightly better result (top-1 validation error rate: 30.88\%) than GN (top-1 validation error rate: 30.96\%), an advanced technique for this scenario with tiny batch size. This promises the correct training of very large model so that each single GPU node can only accommodate several samples.

\section{Conclusion}
Motivated by the importance but high cost of BN layer in modern DNNs, the concept of BN sampling is introduced to simplify the reduction operations. Then two sampling strategies BS and FS are proposed to randomly select a small fraction of data for normalization in the batch dimension and FM dimension, respectively. To balance the requirement for less data correlation and more regular execution graph, an extremely simple variant of BN named VDN is further proposed, which normalizes activations using only one synthetical sample. Experiments on deep networks evidence that the proposed approaches can achieve up to 20\% overall training acceleration with negligible accuracy loss. The accuracy differences among these approaches are analyzed through the visualization of estimation error and estimation correlation. Finally, the VDN can be extended to the micro-BN case with tiny batch size, where it gives comparable accuracy with the state-of-the-art result. This paper preliminary proves the effectiveness and efficiency of BN sampling. All the experiments are conducted on standard Tensorflow library, which is easy-to-use. In the future, developing specialized kernel optimization is promising for more running benefits. For example, the FS pattern at patch grain can be further matched and the padding operation in the backward pass of BN can be removed since our computational graph is fixed during one training epoch.


\bibliography{iclr2019_conference}
\bibliographystyle{iclr2019_conference}

\newpage
\begin{appendices}
\section{Implementation Algorithms}\label{Sec:algorithm}
\textbf{Notations}. We use the convolution layer for illustration, which occupies the major part of most modern networks \citep{he2016deep,Huang2016Densely}. The features can be viewed as a 4D tensor. We use ``$E_{0, 1, 2}$'' and ``$Var_{0, 1, 2}$'' to represent the operations that calculate the means and variances, respectively, where ``{0, 1, 2}'' denotes the dimensions for reduction. 
\\
\begin{algorithm}[H]
\DontPrintSemicolon
 \KwData{input batch at layer $l$: $\displaystyle \tB_l \in \mathbb{R}^{N \times H_l \times W_l \times C_l}$, $l \in$ all layers; sampling size: $n_s \in (0, N]$}
 \KwResult{estimation of $E[\displaystyle \tB_l]$ \& $Var[\displaystyle \tB_l]$, $l \in$ all layers }
 \Begin{
 \For{$ep \in$ all epochs}{
 \For{$l \in$ all layers}{
 if BS: $begin_l=randint(0, N-n_s)$; else NS: $begin_l=0$
 }
 \For{$it \in$ all iterations}{
 \For{$l \in$ all layers}{
 $\displaystyle \tB_s = \displaystyle \tB_l[begin_l: begin_l + n_s-1, :, :, :]$ \\ $E[\displaystyle \tB_l] = E_{0, 1, 2}[\displaystyle \tB_s]$, $Var[\displaystyle \tB_l] = Var_{0, 1, 2}[\displaystyle \tB_s]$\
 }
 }
 }
 }
 \caption{NS/BS Algorithm}
 \label{Alg:NS BS}
\end{algorithm}

\begin{algorithm}[H]
\DontPrintSemicolon
 \KwData{input batch at layer $l$: $\displaystyle \tB_l\in \mathbb{R}^{N \times H_l \times W_l \times C_l}$, $l \in$ all layers; sampling size: $h^{(l)}_s\in (0, H_l]$ \& $w^{(l)}_s\in (0, W_l]$}
 \KwResult{estimation of $E[\displaystyle \tB_l]$ \& $Var[\displaystyle \tB_l]$, $l \in$ all layers}
 \Begin{
 \For{$ep \in$ all epochs}{
 \For{$l \in$ all layers}{
 $begin_h^{(l)}=randint(0, H_l-h^{(l)}_s)$, $begin_w^{(l)}=randint(0, W_l-w^{(l)}_s)$
 }
 \For{$it \in$ all iterations}{
 \For{$l \in$ all layers}{
 $\displaystyle \tB_s = \displaystyle \tB[:, begin_h^{(l)}: begin_h^{(l)} + h^{(l)}_s -1, begin_w^{(l)}: begin_w^{(l)} + w^{(l)}_s -1, :]$ \\ 
  $E[\displaystyle \tB_l] = E_{0, 1, 2}[\displaystyle \tB_s]$, $Var[\displaystyle \tB_l] = Var_{0, 1, 2}[\displaystyle \tB_s]$\
 }
 }
 }
 }
 \caption{FS Algorithm}
 \label{Alg:FS}
\end{algorithm}

\begin{algorithm}[H]
 \KwData{Dataset: $\displaystyle \tD \in \mathbb{R}^{N_D\times H_0 \times W_0 \times C_0}$; input batch: $\displaystyle \tB_l \in \mathbb{R}^{N\times H_l \times W_l \times C_l}$; number of virtual samples: $n_v$}
 \KwResult{estimation of $E[\displaystyle \tB_l]$ \& $Var[\displaystyle \tB_l]$ at layer $l$, $l \in$ all layers}
 \Begin{
 Calculate $\mu = E_{0, 1, 2}[\displaystyle \tD]~\&~\sigma^2 = Var_{0, 1, 2}[\displaystyle \tD]$ offline\\
 \For{$ep \in$ all epochs}{
 \For{$it \in$ all iterations}{
  Create virtual samples $\displaystyle \tV \in \mathbb{R}^{n_v \times H_0 \times W_0 \times C_0}$, $\displaystyle \tV \sim N(\mu, \sigma)$\\
 \For{$l \in$ all layers}{
 if $l=0$: $\displaystyle \tB_l = [\displaystyle \tV, \displaystyle \tB_l]$, then feed $\displaystyle \tB_l$ into the network\\
 $\displaystyle \tB_s = \displaystyle \tB_l[0: n_v-1, :, :, :]$ \\ $E[\displaystyle \tB_l] = E_{0, 1, 2}[\displaystyle \tB_s]$, $Var[\displaystyle \tB_l] = Var_{0, 1, 2}[\displaystyle \tB_s]$\
 }
 }
 }
 }
 \caption{VDN Algorithm}
 \label{Alg:VDN}
\end{algorithm}

\section{Experimental Configuration}\label{Sec:exp config}
All the experiments on CIFAR-10 \& CIFAR-100 are conducted on a single Nvidia Titan XP GPU. We use a weight decay of 0.0002 for all weight layers and all models are trained by 130 epochs. The initial learning rate is set to 0.1 and it is decreased by 10x at 50, 80, 110 epochs. During training, we adopt the "random flip left \& right" and all the input images are randomly cropped to $32\times32$. Each model is trained from scratch for 5 times in order to reduce random variation. 

For ImageNet, We use 2 Nvidia V100 GPUs on DGX station for ResNet-18 and 3 for DenseNet-121. We use a weight decay of 0.0001 for all weight layers and all models are trained by 100 epochs. The initial learning rate is set to $0.1/256\times$``$Gradient~batch~size$'' and we decrease the learning rate by 10x at 30, 60, 80, 90 epochs. During training, all the input images are augmented by random flipping and cropped to $224\times224$. We evaluate the top-1 validation error on the validation set using the centered crop of each image. To reduce random variation, we use the average of last 5 epochs to represent its final error rate. Besides, Wingrad \citep{lavin2016fast} is applied in all models to speedup training.

\section{Analysis of Speedup and Estimation Error}\label{Sec:Speedup analysis}

\subsection{Speedup Analysis}\label{Sec:Cost}

\textbf{Computational Cycle and Memory Access}. Our proposed approaches can effectively speedup forward passes. Under the condition that we sample $s\ll m$ data for each FM, the total accumulation operations are significantly reduced from $m-1$ to $s-1$. If using the adder tree optimization illustrated in Section \ref{sec:Bottleneck}, the tree's depth can be reduce from $log(m)$ to $log(s)$. Thus, the theoretical speedup for the forward pass can reach $log_s(m)$ times. For instance, if the FM size is $56\times 56$ with batch size of 128 ($m=56\times56\times128$), under sampling ratio of 1/32, the speedup will be 36.7\%. The total memory access is reduced by $m/s$ times. For example, when the sampling ratio is 1/32, only 3.1\% data need to be visited. This also contributes a considerable part in the overall speedup.

\textbf{Speedup in the Backward Pass}. The BN operations in the forward pass have been shown in Equation (\ref{equ:BN forward})-(\ref{equ:statistics after sampling}). Based on the derivative chain rule, we can get the corresponding operations in the backward pass as follows 
\begin{equation}
\label{equ:BN backward}
\begin{split}
& \frac{\partial l}{\partial \widehat{x}^{(k)}_i} = \frac{\partial l}{\partial y_i{(k)}} \cdot \gamma^{(k)}, ~~\frac{\partial l}{\partial E[x^{(k)}]} = (\sum_{j=1}^m \frac{\partial l}{\partial \widehat{x}^{(k)}_j} \cdot \frac{-1}{\sqrt{Var[x^{(k)}] + \epsilon}})\\
& \frac{\partial l}{\partial Var[x^{(k)}]} = \sum_{j=1}^m \frac{\partial l}{\partial \widehat{x}_j^{(k)}} \cdot (x_j - E[x^{(k)}]) \cdot \frac{-1}{2}(Var[x^{(k)}] + \epsilon)^{-3/2}\\
& \frac{\partial l}{\partial x_i^{(k)}} = \frac{\partial l}{\partial \widehat{x}_i^{(k)}} \cdot \frac{1}{\sqrt{Var[x^{(k)}] + \epsilon}} + \frac{\partial l}{\partial E[x^{(k)}]} \cdot \frac{1}{m} + \frac{\partial l}{\partial Var[x^{(k)}]} \cdot \frac{2(x_i^{(k)}-E[x^{(k)}])}{m} \\
& \frac{\partial l}{\partial \gamma^{(k)}} = \sum_{j=1}^m \frac{\partial l}{\partial y_i^{(k)}} \cdot \widehat{x}_i^{(k)}, ~~\frac{\partial l}{\partial \beta^{(k)}} = \sum_{j=1}^m \frac{\partial l}{\partial y_i^{(k)}}
\end{split}.
\end{equation}
After BN sampling, the calculation of $\frac{\partial l}{\partial x_i^{(k)}}$ can be modified as
\begin{equation}
\label{equ:BNS backward}
\frac{\partial l}{\partial x_i^{(k)}} = \left\{
\begin{array}{lr}
\frac{\partial l}{\partial \widehat{x}_i^{(k)}} \cdot \frac{1}{\sqrt{Var[x^{(k)}] + 
\epsilon}} + \frac{\partial l}{\partial E[x^{(k)}]} \cdot \frac{1}{s} + \frac{\partial l}{\partial Var[x^{(k)}]} \cdot \frac{2(x_i^{(k)}-E[x^{(k)}])}{s},~ \text{if}~x_i \in S \\
\frac{\partial l}{\partial \widehat{x}_i^{(k)}} \cdot \frac{1}{\sqrt{Var[x^{(k)}] + 
\epsilon}},~\text{otherwise}
\end{array},
\right.
\end{equation}
while the others remain the same. Here $S$ is the location set for sampled neurons. For neurons outside $S$, they didn't participate in the estimation of the mean and variance in the forward pass, so $\frac{\partial E[x^{(k)}]}{\partial x_i^{(k)}}$ and $\frac{\partial Var[x^{(k)}]}{\partial x_i^{(k)}}$ equal to zeros. We should note that although we sampled only $s$ data in the forward pass, the reduction operations in the backward pass still have $m$ length since the neurons outside $S$ also participate in the activation normalization using the estimated statistics from the sampled data. Therefore, there is no theoretical speedup for the backward pass. In our experiments, to assemble a dense and efficient addition operation with $\frac{\partial l}{\partial \widehat{x}_i^{(k)}} \cdot \frac{1}{\sqrt{Var[x^{(k)}] + 
\epsilon}}$ for all neurons, the mentioned zeros outside $S$ are pre-padded. However, it has great potential for memory space and access reduction since the $\frac{\partial E[x^{(k)}]}{\partial x_i^{(k)}}$ and $\frac{\partial Var[x^{(k)}]}{\partial x_i^{(k)}}$ outside $S$ are zeros, which can be leveraged in specialized devices.

\textbf{Regular Sampling Pattern and Static Sampling Index}. To approach the theoretical speedup, we suggest to use more regular sampling pattern such as the continuous samples in BS and intra-FM rectangular shape and inter-FM shared location in FS. A regular sampling pattern can improve the cache usage by avoiding random access. Moreover, the operation in the backward pass corresponding to the sampling operation in the forward pass is ``padding" shown in Fig. \ref{fig:Bottleneck}(d). Under a regular sampling pattern, the padding is block-wise that becomes much easier. On the other side, intuitively, a static computation graph can be calculated faster and easier deployed on various platforms for (1) the static computing graphs' pipeline can be optimized once for all before training, (2) popular deep learning frameworks like TensorFlow \citep{girija2016tensorflow} are developed as a static computation graph. For these reasons, we expect the random sampling is achieved through a static computational graph, which is obtained by updating the sampling indexes only per epoch.

\subsection{Estimation Error Analysis}\label{Sec:Error}

\textbf{Intra-layer Data Correlation}. For each layer, the variance of estimated statistics (e.g. mean) is highly related to the correlation between sampled data, which can be described by
\begin{equation}
\label{equ:intra-layer estimation}
Var[E[x_s^{(k)}]]= Var[\frac{1}{s}\sum_{j=1}^sx_j^{(k)}] = \frac{1}{s^2}(\sum_{i=1}^s Var[x_i^{(k)}] + 2\sum_{i \ne j}^s Cov(x_i^{(k)}, x_j^{(k)})).
\end{equation}
Obviously, less data correlation can reduce the estimation variance, thus bring more accurate estimation, which illustrates why FS produces better accuracy than BS: the selected data come from all samples which are more uncorrelated.

\textbf{Inter-layer Data Correlation}. \cite{bjorck2018understanding} suggested that BN's effectiveness is probably because it can prevent the input statistics at each layer from scaling up. Under BN sampling, if positively correlated data are selected at each layer (e.g. sharing the same sampling indexes in each layer), the estimated statistics in different layers are also probably correlated. This may lead to correlated estimation error between layers which would further result in exponentially increasing estimation error throughout the network layer by layer. Oppositely, if we select less correlated data at each layer, the effectiveness of original BN can be preserved better.

\section{Influence of Decay Rate for Moving Average}\label{Sec:decay rate} 

During each validation step after certain training iterations, $E[x^{(k)}]$ \& $Var[x^{(k)}]$ are replaced with the recorded moving average, which is governed by 
\begin{equation}
\label{equ:moving average1}
X_{ma}[x^{(k)}]_{it} = \left\{
\begin{array}{lr}
X[x^{(k)}]_{it}, {it} = 1\\
\alpha X[x^{(k)}]_{it} + (1 - \alpha ) X_{ma}[x^{(k)}]_{it-1}, it \in [2, \frac{N}{B}]
\end{array},
\right.
\end{equation}
where $X$ stands for ``$E$'' or ``$Var$'', $X_{ma}[x^{(k)}]$ denotes the moving average, $it$ is the iteration number, and $\alpha$ is the decay rate. Based on Equation (\ref{equ:moving average1}), which the variance of $X_{ma}[x^{(k)}]_{it}$ equals to $\alpha^2 Var[X[x^{(k)}]_{it}] + (1 - \alpha )^2 Var[X_{ma}[x^{(k)}]_{it-1}]$, if we assume that (1) each estimated value shares the same variance $ Var[X[x^{(k)}]_{it}]$ and (2) independent with each other while (3) the iteration number goes larger, we will get \eqref{equ:moving average2}.
\begin{equation}
\label{equ:moving average2}
Var[X_{ma}[x^{(k)}]_{it}]\approx \frac{\alpha}{2-\alpha} Var[X[x^{(k)}]_{it}].
\end{equation}

\begin{wrapfigure}{r}{0.35\textwidth}
\includegraphics[width=0.35\textwidth]{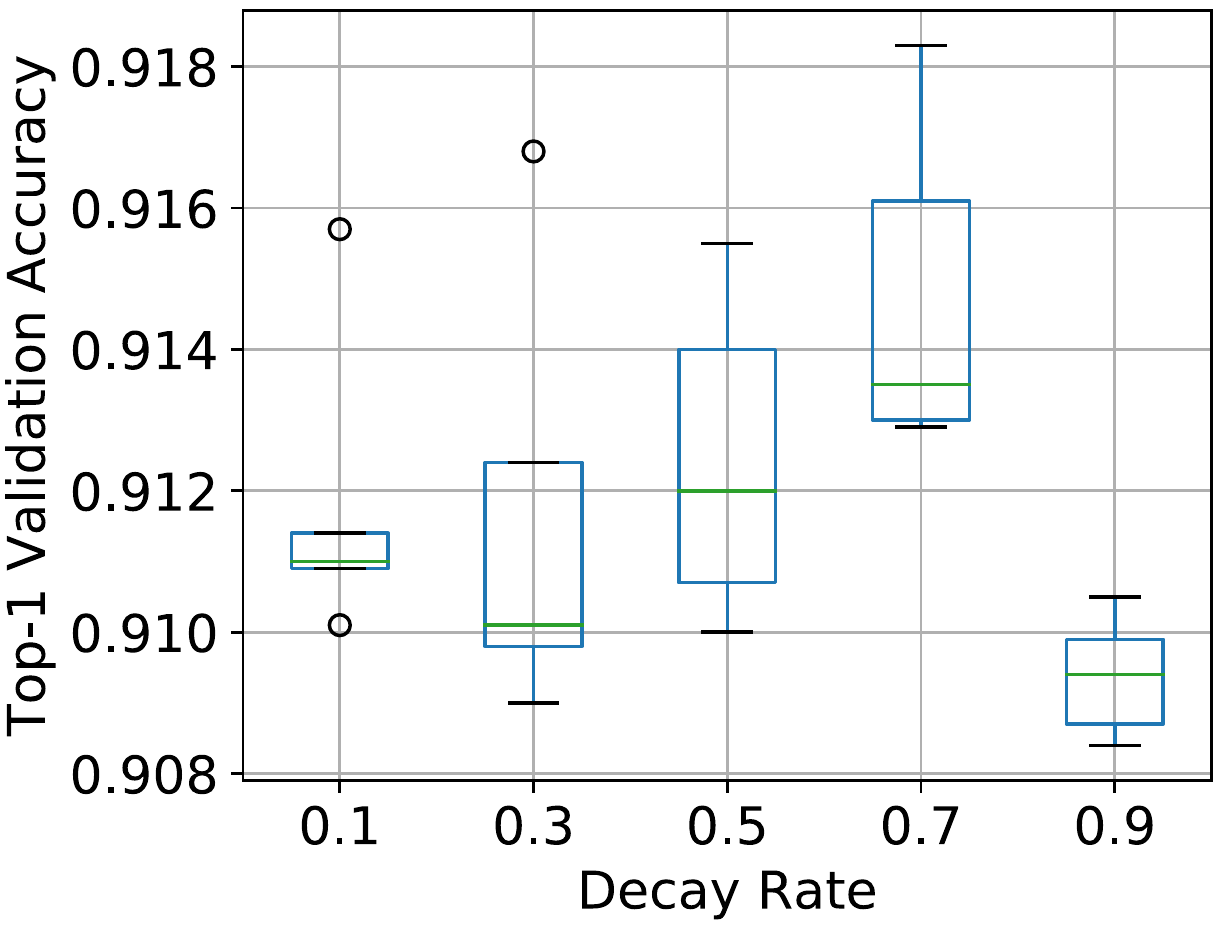}
\caption{Influence of decay rate for moving average.}
\label{fig:decay rate}
\end{wrapfigure}

The above equation reveals that an appropriately smaller $\alpha$ might scales down the estimation error, thus produces better validation accuracy. To verify this prediction, the experiments are conducted on ResNet-56 over CIFAR-10 and using BS-1/128(0.78\%) sampling. As shown in Fig. \ref{fig:decay rate}, it's obvious that there exists a best decay rate setting (here is 0.7) whereas the popular decay rate is 0.9. The performance also decays when decay rate is smaller than 0.7, which may because a too small $\alpha$ will lose the capability to record the moving estimation, thus degrade the validation accuracy. This is interesting because the decay rate is usually ignored by researchers, but the default value might be not the best setting for BN sampling.

\end{appendices}

\end{document}